\documentclass{article}

\usepackage{arxiv}

\usepackage[utf8]{inputenc} 
\usepackage[T1]{fontenc}    
\usepackage{hyperref}       
\usepackage{url}            
\usepackage{booktabs}       
\usepackage{amsfonts}       
\usepackage{nicefrac}       
\usepackage{microtype}      
\usepackage{lipsum}
\usepackage{graphicx}
\graphicspath{ {./images/} }
\usepackage{amsmath}
\usepackage{caption}
\usepackage{subcaption}
\usepackage{url}            
\usepackage{amsfonts}       
\usepackage{nicefrac}       
\usepackage{multirow}
\usepackage{float}
\usepackage{lscape}
\usepackage[paper=portrait,pagesize]{typearea}
\usepackage{geometry}
\usepackage{lineno}

\title{Turning old models fashion again: Recycling classical CNN networks using the Lattice Transformation}

\author{
 Ana Paula G. S. de Almeida \\
  Department of Mechanical Engineering \\
  University of Brasilia\\
  Brasilia, DF, Brazil \\
  \texttt{anapaula.gsa@gmail.com} \\
   \And
 Flavio de Barros Vidal \\
  Department of Computer Science\\
  University of Brasilia\\
  Brasilia, DF, Brazil. \\
  \texttt{fbvidal@unb.br} \\
}

\begin{document}
\maketitle
\begin{abstract}
In the early 1990s, the first signs of life of the CNN era were given: LeCun {\it et al.}~\cite{lecun1990handwritten} proposed a CNN model trained by the backpropagation algorithm to classify low-resolution images of handwritten digits. Undoubtedly, it was a breakthrough in the field of computer vision. But with the rise of other classification methods, it fell out fashion. That was until 2012, when Krizhevsky {\it et al.} (2012)~\cite{alexnet} revived the interest in CNNs by exhibiting considerably higher image classification accuracy on the ImageNet challenge. Since then, the complexity of the architectures are exponentially increasing and many structures are rapidly becoming obsolete. Using multistream networks as a base and the feature infusion precept, we explore the proposed LCNN~\cite{lattice} cross-fusion strategy to use the backbones of former state-of-the-art networks on image classification in order to discover if the technique is able to put these designs back in the game. In this paper, we showed that we can obtain an increase of accuracy up to $63.21\%$ on the NORB dataset we comparing with the original structure. However, no technique is definitive. While our goal is to try to reuse previous state-of-the-art architectures with few modifications, we also expose the disadvantages of our explored strategy.
\end{abstract}


\section{Introduction}

Colorado, 1989. In the Neural Information Processing Systems conference, a game-changer approach of image classification was presented by LeCun {\it et al.}~\cite{lecun1990handwritten}. A breakthrough in the field of computer vision was made when they used Convolutional Neural Networks trained by backpropagation to categorize low-resolution images of handwritten digits. Thenceforward, the resolution of high-level problems scaled expeditiously; the area of computer vision had reached a new and elaborate level. LeNet-1 --- a multi-layer network with convolutions with small size kernels followed by a squashing function and additional local averaging and subsampling layers, diminishing the resolution of the generated feature map --- was just the beginning, and the evolution continues to this day. In 1998, LeCun~\cite{726791} published a paper reviewing miscellaneous approaches applied to handwritten character recognition and compared them, showing that CNNs outperform all other techniques. It also proposes an evolution to the original LeNet-1, LeNet-5. Nonetheless, the simplicity of the architectures, directly tied to the hardware of the time, was short-lived.

According to Girshick {\it et al.} (2014)~\cite{girshick2014rich}, CNNs saw massive usage in the 1990s~\cite{lecun1990handwritten,726791}, but then fell out of fashion with the rise of Support Vector Machines (SVMs). In 2012, however, Krizhevsky {\it et al.} (2012)~\cite{alexnet} revived the interest in CNNs by exhibiting considerably higher image classification accuracy --- $84.7\%$ --- on the ImageNet Large Scale Visual Recognition Challenge (ILSVRC)~\cite{imagenet}, almost halve the error rate for object recognition at the time. Likewise, the increasing hardware processing capability and popularization of Graphical Processing Units (GPUs) were additional assets to the works.

\begin{figure}[htpb]
	\centering
	\includegraphics[scale=0.5]{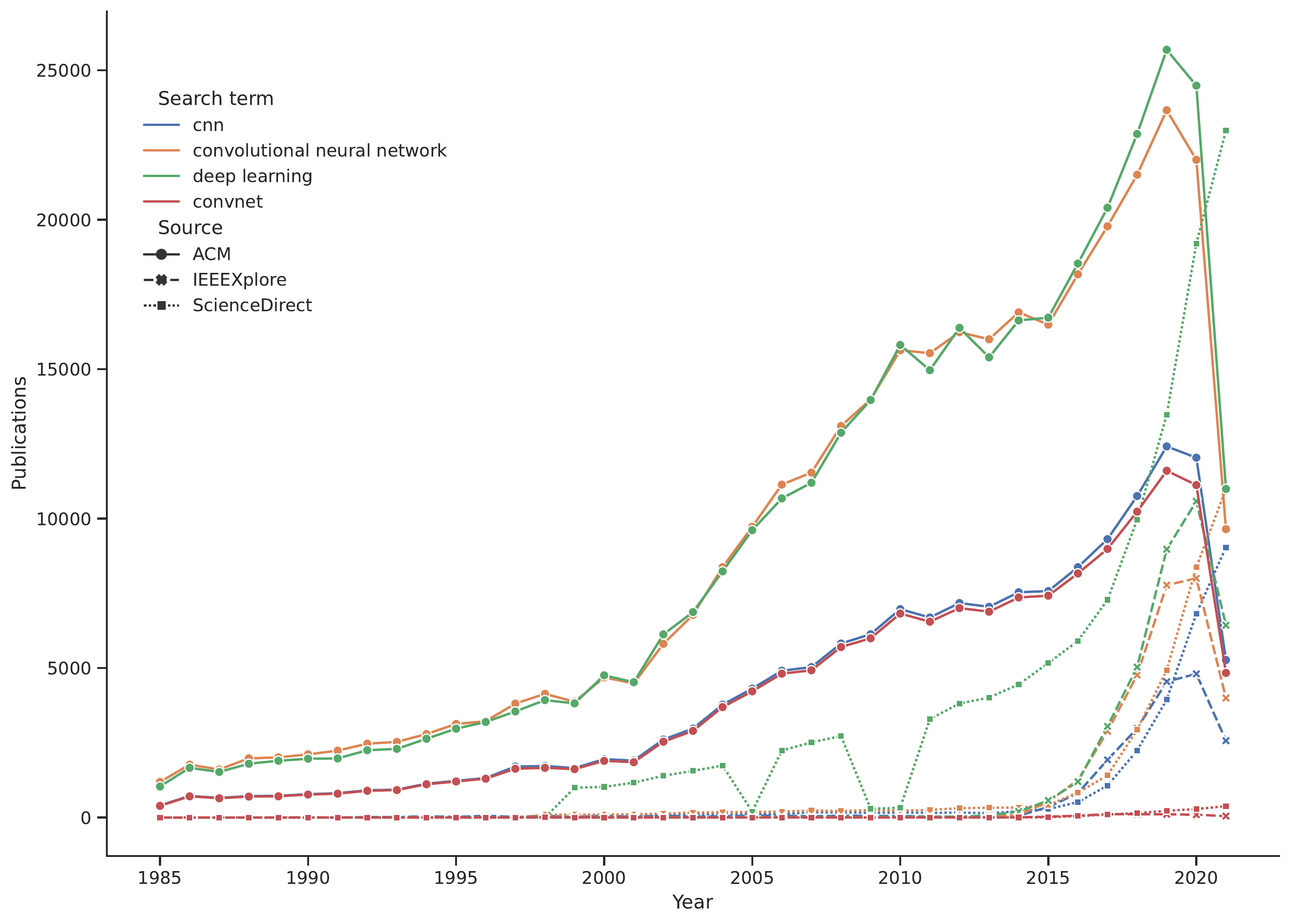}
	\caption{Occurrences of the terms: \textit{CNN, convolutional neural network, ConvNet} and \textit{deep learning} in 3 different repositories (ACM~\cite{acm}, Elsevier (ScienceDirect)~\cite{sciencedirect} and IEEEXplore~\cite{ieeexplore}).}
	\label{chap1:cnn_occur}
\end{figure}

The impact of the convolutional network called Alexnet and the revival of the CNN topic on images can also be seen in the graphic presented in Figure~\ref{chap1:cnn_occur} of publications by year, since 1985, from three of the largest digital research libraries of the Artificial Intelligence field: ACM, IEEE Xplore, and ScienceDirect. Correlate terms as \textit{convolutional neural network, ConvNet} and \textit{deep learning} were also used to highlight the effects of such event.

Figure~\ref{chap1:cnn_occur} also shows that brief research is enough to encounter a variety of state-of-the-art winning architectures. But are the early designs so out-of-date and unusable that we have to keep modifying them with such frequency?

Besides complex architecture creations, new approaches appeared using more than one of the already established structures to solve classification problems, taking advantage of the infusion of more features in the model: Multistream, or Multichannel, Convolutional Neural Networks (MCNN) have been refined and used in many distinct applications~\cite{karpathy, abade, rw_gammulle2017two, feichtenhofer2016convolutional, 8219720, 8614102, DBLP:journals/tcsv/TuXDLY19, 8513556}. Originally derivative from traditional CNNs, this kind of model architecture allows to employ nearly all, conventional (or not) models, available in the literature, as \textit{LeNet}\cite{726791}, \textit{Alexnet}\cite{alexnet}, \textit{ResNet}\cite{resnet} and many others, by basically adjusting (or modifying) the fusion stage\cite{karpathy}. Normally, multistream approaches are an alternative to multimodal tasks, such as video classification or gesture recognition, due to their capability of feeding the network with an extra set of features. 

In Karpathy \textit{et al.} (2014)~\cite{karpathy}, the fusion issue was taken to a whole new level: two individual single-frame networks were arranged time-delayed apart and their outputs were merged in a fusion procedure, improving the accuracy scores on video action recognition applications. Feichtenhofer \textit{et al.} (2014)~\cite{feichtenhofer2016convolutional} went a little further on network fusions, showing that this additional step can be implemented at a convolutional layer sans loss of accuracy. 

A model based on a cross-fusion method applied for image classification multistream approaches, the Lattice-Convolutional Neural Network (LCNN), was presented in Almeida \textit{et al.} (2019)~\cite{lattice}. This general architecture, that can be adapted for any CNN with an ReLU activation output, is focused on the fusion stages of the network. Using input distractors to prove the efficacy of the technique, the cross-fusion strategy is focused on observed features, particularly in the ReLU's block, used by many classical MCNNs models developed in several popular approaches (e.g. in \cite{karpathy,Velickovic_2016,feichtenhofer2016convolutional}).

Since the LCNN model targets how the fusions are performed can improve the final model accuracy, in this paper, we explored more of this technique by expanding the tests of the aforementioned strategy, gathering more datasets for a fair comparison, and adapting eight traditional CNN models of different shapes: Alexnet~\cite{alexnet}, Xception~\cite{chollet2017xception}, ResNet-18, ResNet-34, ResNet-50~\cite{resnet}, DenseNet-121, DenseNet-169 and DenseNet-201~\cite{densenet}. All these architectures were, at a given time, state-of-the-art on ImageNet competition~\cite{imagenet}, a reference on the image classification task, and they fill the LCNN main requirement: ReLU activations. Here, we show that a slight modification can bring these once state-of-the-art back to the competition.

We reaffirm that our intention is not to achieve the state-of-the-art accuracy, but rather to indicate that models that are no longer considered the best in performance can receive an extra life. This recycling process can provide power to out-of-date models, reusing a backbone that has already been developed.

Using CIFAR-10 dataset as reference, our lattice strategy outperforms a simple late fusion of the models in at most $30.78\%$ --- L-ResNet-50 with average operation --- and at least $0.54\%$ --- L-DenseNet-121 with addition operation ---.

Furthermore, it also presents an adaptability when it comes to testing with different types of mathematical operations, providing an auxiliary alternative when results are not so good.

Thus, this paper is divided as follows: Section~\ref{relatedworks} presents a brief background on multiple stream networks, our primary basis. Section~\ref{lattice} introduces the L-strategy module, while Sections~\ref{experiment} and~\ref{results} shows our experiments and discussions, with positive and negative aspects of the technique. Section~\ref{conclusion} concludes the work with its future projections.


\section{Background}
\label{relatedworks}

\subsection{Convolutional Neural Networks}
The late 90's LeNet-5~\cite{726791} is a 7-layer CNN that receives as input an approximately size-normalized and centered 32x32 pixel image. This input size was meant to be larger than the images of the dataset in order to center the potential distinctive features of the data, such as corners. The first convolutional layer, C1, has six 28x28 sized feature maps. Each unit in each feature map is connected to a 5x5 neighborhood in the input. Afterward, there is a subsampling layer with six 14x14 sized feature maps. These feature maps' units are connected to a 2x2 neighborhood in the corresponding feature map in C1.

These connections are sustained until the third convolutional layer. This behavior causes a break of symmetry in the network, so different feature maps are forced to extract different features~\cite{726791}.

Following the breakthrough models, the AlexNet architecture~\cite{alexnet} has eight layers. With an input size of $224 \times 224 \times 3$, the first convolutional layer filters the image with 96 kernels of size $11 \times 11 \times 3$ with a distance between the receptive field centers of neighboring neurons in a kernel map of 4 pixels. The output of the first convolutional layer becomes the input of the second convolutional layer. The third, fourth, and fifth convolutional layers are connected without pooling layers. The first two fully-connected layers use a dropout~\cite{dropout} technique to reduce data overfitting.

ResNet~\cite{resnet} is also a deep architecture that relies in modules to prevent a learning degradation with the increase of depth. Instead of learning a direct mapping of $x \rightarrow y$ with a few stacked non-linear layers ($H(x)$), a residual mapping is proposed, where $F(x) := H(x) — x$, which can be reframed into $H(x) = F(x) + x$, where $F(x)$ and $x$ represents the stacked non-linear layers and the identity function respectively.

According to He {\it et al.} (2015)~\cite{resnet}, it is easier to optimize the residual mapping than the unreferenced mapping. Likewise, the formulation of $H(x) = F(x) + x$ can be realized by feedforward neural networks with shortcut connections. In the \textit{ResNet Module}, these shortcut connections perform identity mapping, and their outputs are added to the outputs of the stacked layers. To evaluate the explored strategy herein, we used ResNet-18, ResNet-34 and ResNet-50.

Xception~\cite{chollet2017xception} network is an interpretation of Inception modules~\cite{googlenet} with linear stacks of depthwise separable convolution layers with residual connections. Moreover, in 2018, DenseNets or Densely Connected Convolutional Networks~\cite{densenet} were introduced. They are an architecture that connects each layer to every other layer in a feed-forward way so maximum information flow between layers in the network is guaranteed. In this model, for each layer, the feature-maps of all earlier layers are used as inputs, and its own feature-maps are used as inputs
into all following layers. Here, we modified architectures DenseNet-121, DenseNet-169 and DenseNet-201.

\subsection{Multistream Convolutional Neural Networks}
Karpathy  {\it et al.} (2014)~\cite{karpathy} proposed an empirical evaluation of CNNs on video classification based on the fact that the standard video classification approaches~\cite{dollar2005behavior, laptev2005space, laptev2008learning, 5206744} consisted in involving visual features extraction, feature combination, and then classification and also that CNNs were achieving state-of-the-art results on different fields of computer vision~\cite{alexnet, seg2, obj1}, a new technique that included local motion information present in the video as connectivity to a CNN architecture was suggested.

\begin{figure*}[!htpb]
\centering
\includegraphics[width=.9\textwidth]{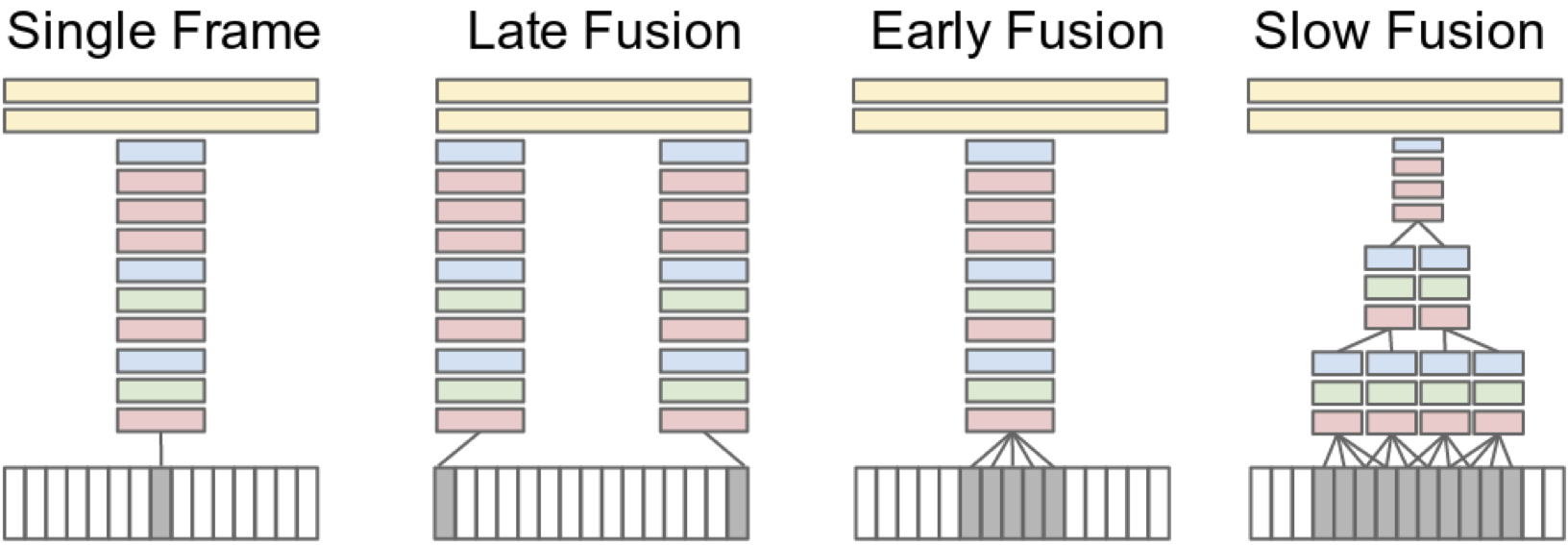}
\caption{Fusion stages presented by Karpathy  {\it et al.}(2014)~\cite{karpathy}.  Figure originally found in~\cite{karpathy}.}
\label{karpathy}
\end{figure*}

This technique is a multiresolution architecture that extends the connectivity of a CNN and uses the additional information as a way to improve overall performance. Using an Alexnet~\cite{alexnet} as a baseline, Karpathy  {\it et al.} explore three different types of network fusion: early, late, and slow. These fusion stages, illustrated in Figure~\ref{karpathy}, are concatenations in between layers and late fusion is the preferred method to this day.

In the same year, Simonyan \textit{et al.} (2014) proposed a two-stream architecture for the purpose of receiving spatial and temporal components of a video. However, in this work, two fusion methods are considered: averaging and training a multi-class linear SVM on stacked $L_{2}$-normalized softmax scores as features. Inspired by this proposal, Feichtenhofer {\it et al.} (2016)~\cite{feichtenhofer2016convolutional} propose a fusion evaluation of CNNs, to best take advantage of the additional information that this kind of network disposes of. Using the two-stream architecture previously proposed by Simonyan {\it et al.} (2014)~\cite{simonyan2014two}, the authors consider different types of fusion, for spatial --- sum, maximum, concatenation, convolution, and bilinear --- and temporal --- 3D pooling and 3D convolution + pooling --- features.

\begin{figure*}[!htpb]
\centering
\includegraphics[width=.85\textwidth]{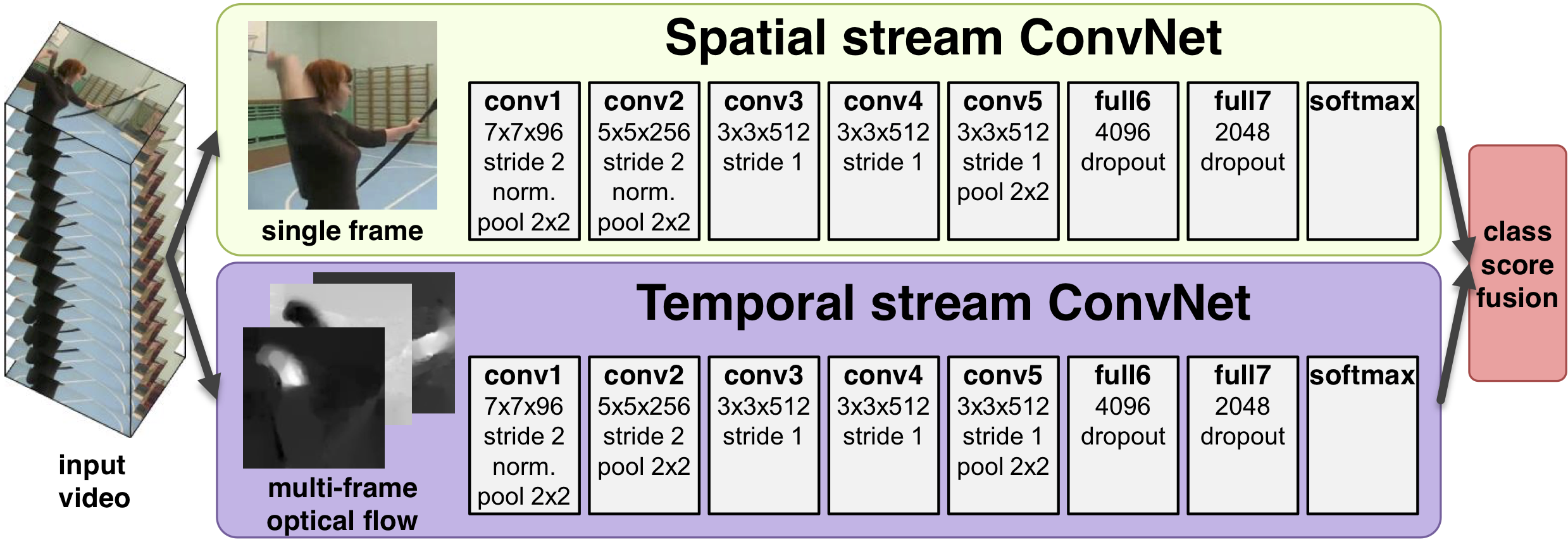}
\caption{Two-stream architecture.  Figure originally found in~\cite{simonyan2014two}.}
\label{simonyan}
\end{figure*}

It is still unusual to work with multiple streams in an image classification context. Aerial images, for example, have not only space and texture features but also contain a large number of scene semantic information~\cite{rw_yu2018two}. Thus, a good feature representation is needed for positive classification results. Yu {\it et al.} (2018)~\cite{rw_yu2018two} proposed a two-stream CNN based on two pre-trained CNNs as feature extractors to learn deep features from the original aerial image and the processed aerial image through saliency detection, respectively. Right before an extreme learning machine classifier, the streams are fused in two distinct strategies -- concatenation and sum.

Abade {\it et al.} (2019)~\cite{abade} use a multistream approach in order to diagnose plant diseases. It adapts classical CNN models to train and evaluates the PlantVillage dataset. The adaptation was made so the different versions of the dataset could be used. The fusion strategy, however, is a simple late fusion.

As previously shown, multiple stream networks have been developed, applied, and used in many situations and applications nowadays~\cite{rw_gammulle2017two, 8219720, 8614102, DBLP:journals/tcsv/TuXDLY19, 8513556, 8371385}. Nevertheless, signal treating in between multistream layers strategies lack in literature. Velickovic \textit{et al.} (2016)~\cite{Velickovic_2016} present a cross-modal architecture for image classification, the X-CNN.

\begin{figure*}[!htpb]
\centering
\includegraphics[width=\textwidth]{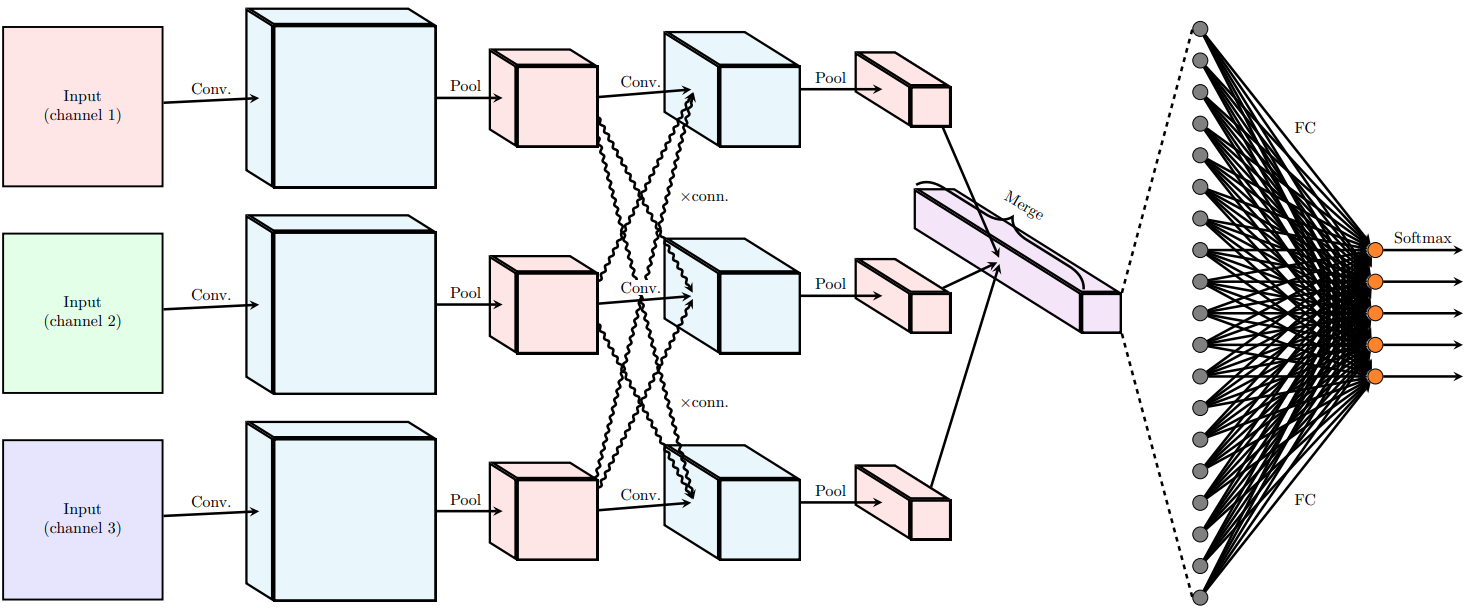}
\caption{X-CNN architecture.  Figure originally found in~\cite{Velickovic_2016}.}
\label{xcnn}
\end{figure*}

Designed to deal with sparse data sets, X-CNNs are a typically image-based approach that allows weight-sharing and information exchange between hidden layers of a network by using cross-connections inserted after each pooling layer, presented in Figure~\ref{xcnn}. Inspired by the biological cross-connections between various sensory networks, this strategy attempts to improve CNNs predictions without requiring large amounts of input data dimensionality.

Akilan \textit{et al.} (2017)~\cite{8122666} explore late fusion upon different multi-deep CNNs used as feature extractors. Their approach uses a four rule feature fusion -- product, sum, average and maximum -- in order to merge different CNN features. Based on previous works that ensembled distinct classifiers, such as K-Nearest Neighbors (KNN) and CNNs, they demonstrate once more that even simple fusion processes can improve classification results.

Still following the line of multiple CNN architectures ensemble, Amin-Naji \textit{et al.} (2019)~\cite{AMINNAJI2019201} also use several networks simultaneously trained on a dataset to solve an issue. This time, however, the proposed methodology also intends to create multiple focuses on the dataset, improving the overall accuracy. For fusing the networks, a concatenation strategy is used.

As said before, it is typical to observe multiple streams when treating videos, that have spatial and temporal portions, or when using different data modalities. In~\cite{Velikovi2018}, multimodal time-series data is analyzed using an X-LSTM technique. Each input passes through a separate three-layer LSTM stream, allowing a piece of information to flow using cross-connections between the streams in the second layer, where features from a stream are passed and concatenated with features from another stream. In~\cite{perez}, they propose a search space that covers numerous possible fusion architectures given the nature of the multimodal dataset. The outputs of layers that perform a function, such as convolutions or poolings, are then eligible to be chosen for fusion in this approach. When a fusion point is selected, a concatenation operation is performed. The XFlow network~\cite{8894404} also brings an ensemble of different architectures, but with cross-connections as fusing strategies.

\begin{figure}[!htpb]
\centering
\includegraphics[width=0.4\textwidth]{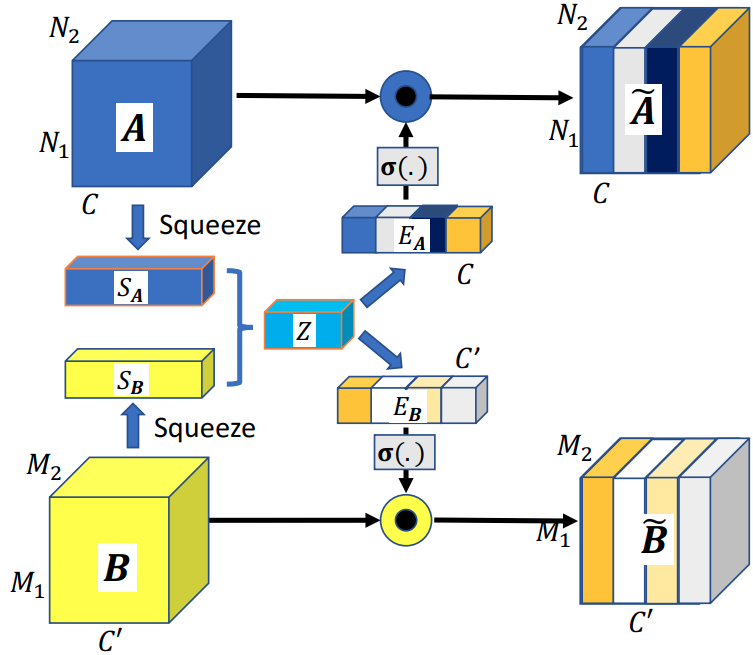}
\caption{MTMM module. Figure originally found in~\cite{joze2020mmtm}.}
\label{mtmm}
\end{figure}

In 2020, Joze \textit{et al.}~\cite{joze2020mmtm} proposed a multimodal transfer module for CNN fusions -- MTMM --. The MTMM can be added at different levels of the model and one main advantage is that the input tensors do not have to have the same spatial dimensions, as it performs squeeze and excitation operations. Also, each stream can be initialized with existing pre-trained weights, since minimum chances are made in the main structure. Although the module is not mainly used for image classification, it can be adapted to this task. An illustration of the module can be found in Figure~\ref{mtmm}, where $A$ and $B$ are layer features.

\section{Lattice Cross-fusion Strategy}
\label{lattice}
In system analysis, there is a field that studies the enhancement or restoration of degraded signals~\cite{signals_oppenheim}. Inspired by the fact that there is no signal processing without degradation and that a CNN signal suffers loss along its course, we decided to apply basic signal operations in between activation layers. Especially the ones that use the ReLU (Rectified Linear Unit) function: $g(z) = max\{0, z\}$, where $g(z)$ is a non-linear function; used by many classical MCNNs models developed in several popular approaches (e.g. in~\cite{karpathy,Velickovic_2016,feichtenhofer2016convolutional}). ReLU outputs zero across half its domain, making the derivatives through it remain significant whenever the unit is active. Nonetheless, they cannot learn with gradients near zero~\cite{deep_learning_goodfellow}.

In order to boost up near zero gradients and get a hold of important features that may be left out, we combine two different signal streams and in a crossing signal inference of the operation result to the input of the subsequent layer. Equation~\ref{eq_fusion} presents a fusion $F$ of two signals $s, \bar{s}$, where $\odot$ represents the chosen mathematical operation.

\begin{equation}
\centering
    F(s,\bar{s}) = s \odot \bar{s}.
\label{eq_fusion}
\end{equation}

Then, a layer $L$ can be defined as:

\begin{equation}
\centering
\begin{aligned}
     L(s,\bar{s}) =[F_{a}(C_{a}(s), C_{b}(\bar{s})), \\
     F_{b}(C_{b}(\bar{s}), C_{b}(s))].
\end{aligned}
\label{lat_eq1}
\end{equation}

As described in Almeida \textit{et al.} (2019)~\cite{lattice}, the cross-fusion function $F: C_{a}, C_{b},..., \\C_{k},... C_{n} \rightarrow y$ combines the $n-1$ convolutional layers with the $C_{k} \in \mathbb{R}^{h \times w \times d}$  where  $h, w$ and $d$ are the height, width, and depth (number of channels/streams), respectively.

The cross-fusion general module is defined in Equation~\ref{lat_eq1}. It computes the operation $\odot$ of two convolutional layers inputs, $C_{a}$ and $C_{b}$, connecting the result as an input of the next layers $C_{a'}$ and $C_{b'}$. It is important to point out that any mathematical operation available in the Tensorflow library~\cite{abadi2016tensorflow} can be applied in the $\odot$ stage, such as addition, subtraction, average.

\begin{figure*}[!htpb]
  \centering
  \includegraphics[width=0.75\textwidth]{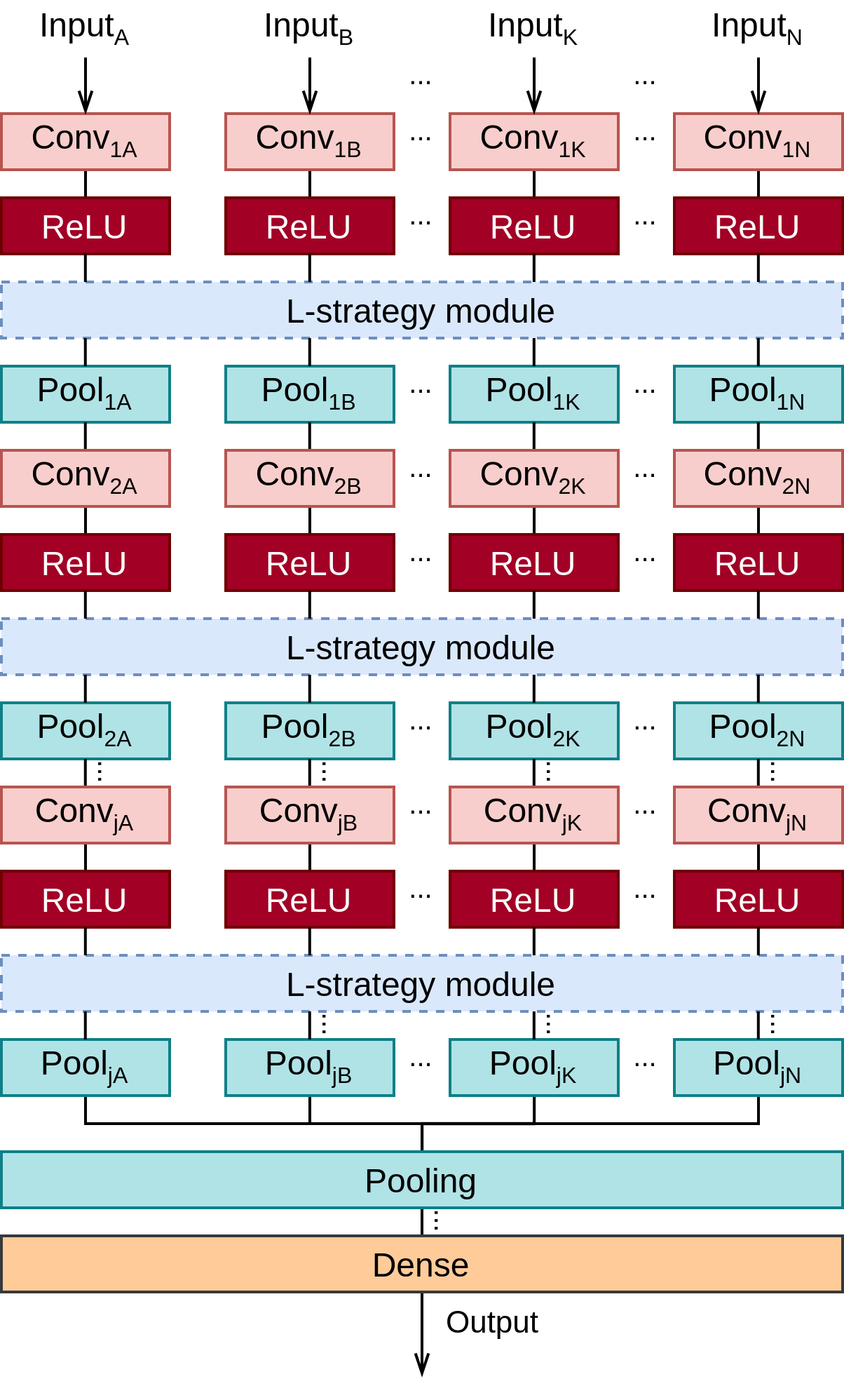}
  \caption{a general L-CNN model. Convolutional layers are represented by the color red, followed by a wine color ReLU indicator, while the fusion modules are dotted-blue. Poling layers are expressed by the color green. Other layers are included for an architecture disclosure.}
  \label{generall}
\end{figure*}

These fusion modules are repeated along with all CNNs' \textit{convolution-ReLU} layers sets, finishing with a late fusion process right before the fully connected stack step. Figure~\ref{generall} presents a visual definition of this proposed cross-function strategy.

\begin{figure}[H]
  \centering
  \includegraphics[width=0.45\textwidth]{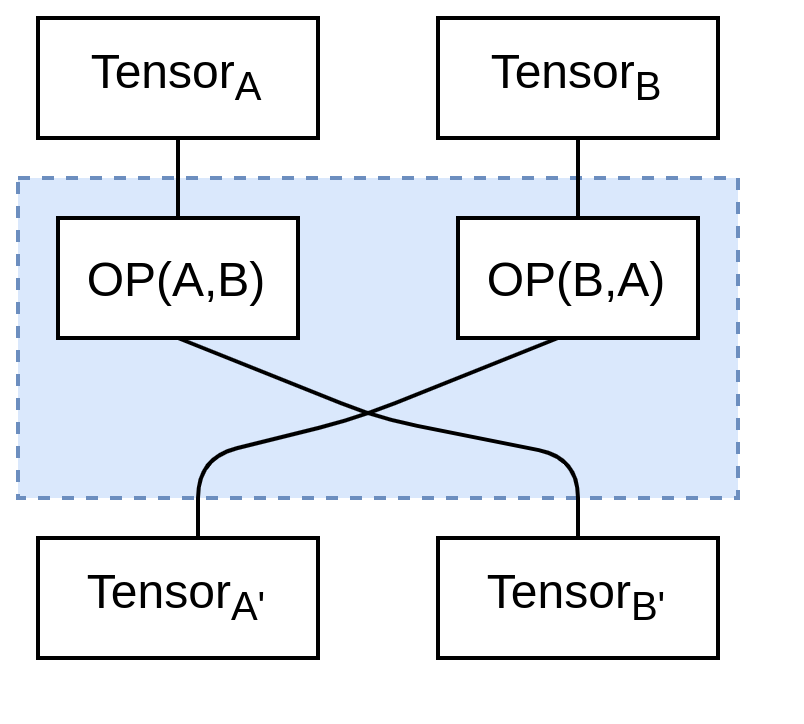}
  \caption{A closer look at the L-strategy module. $OP$ represents a mathematical operation.}
  \label{lmodule}
\end{figure}

The signal crossing can be noticed in the visualization of the L-strategy module found in Figure~\ref{lmodule}.

With this signal enhancement, it is possible to empower CNNs with fewer layers and use simpler structures to achieve as good results as state-of-the-art networks that have tons of convolutional layers, modules and cannot run in modest hardware. The next Section shows our work methodology and we demonstrate that our results, even with insignificant inputs, generally outperform classical late fusion approaches.

\section{Experimental evaluation}
\label{experiment}
Knowing that the addition of features in another stream will naturally improve the network accuracy, we want to demonstrate that the lattice strategy has significant improvements even with distractors or non-quality features being included on the input of the network. Since our cross-fusion strategy can be applied to any network that has convolutions with ReLU activations, we chose to evaluate our method on eight distinct architectures, varying in depth, the number of convolutional layers and parameters in general. Further, three fusion operations were used.

\subsection{Network architectures}
For our experimental evaluation, 8 different network architectures were used as backbones to new networks that used the cross-fusion module, all of them detailed in Section~\ref{relatedworks}. Except for AlexNet, the other architectures are very large and we are not going to detail our L-strategy blocks. Three mathematical operations were implemented in our fusion module: average, addition and subtraction. Full models can be found in the appendix section. All the experiments were made using a default SGD optimizer with a 0.01 learning rate.

\begin{table*}[!htpb]
\centering
\begin{tabular}{c|c|c}
\textbf{Architecture}         & \textbf{Type}             & \textbf{Parameters} \\ \hline\hline
\multirow{3}{*}{Alexnet}      & Single stream             & 23981450            \\ \cline{2-3} 
                              & Multistream - late fusion & 30087946            \\ \cline{2-3} 
                              & Multistream - lattice     & 30087946            \\ \hline
\multirow{3}{*}{ResNet-18}    & Single stream             & 11192458            \\ \cline{2-3} 
                              & Multistream - late fusion & 22384906            \\ \cline{2-3} 
                              & Multistream - lattice     & 34937866            \\ \hline
\multirow{3}{*}{ResNet-34}    & Single stream             & 21311754            \\ \cline{2-3} 
                              & Multistream - late fusion & 42623498            \\ \cline{2-3} 
                              & Multistream - lattice     & 65295754            \\ \hline
\multirow{3}{*}{ResNet-50}    & Single stream             & 23592842            \\ \cline{2-3} 
                              & Multistream - late fusion & 47185674            \\ \cline{2-3} 
                              & Multistream - lattice     & 57338634            \\ \hline
\multirow{3}{*}{DenseNet-121} & Single stream             & 7047754             \\ \cline{2-3} 
                              & Multistream - late fusion & 14095498            \\ \cline{2-3} 
                              & Multistream - lattice     & 14087306            \\ \hline
\multirow{3}{*}{DenseNet-169} & Single stream             & 12659530            \\ \cline{2-3} 
                              & Multistream - late fusion & 25319050            \\ \cline{2-3} 
                              & Multistream - lattice     & 25305738            \\ \hline
\multirow{3}{*}{DenseNet-201} & Single stream             & 18341194            \\ \cline{2-3} 
                              & Multistream - late fusion & 36682378            \\ \cline{2-3} 
                              & Multistream - lattice     & 36667018            \\ \hline
\multirow{3}{*}{Xception}     & Single stream             & 20881970            \\ \cline{2-3} 
                              & Multistream - late fusion & 41763930            \\ \cline{2-3} 
                              & Multistream - lattice     & 41728538            \\
\end{tabular}                             
\caption{Network parameters by implementation type.}
\label{parameter}
\end{table*}

\subsection{Data}
\begin{figure}[!htpb]
    \centering
    \begin{subfigure}[t]{.35\textwidth}
      \centering
      \includegraphics[width=\linewidth]{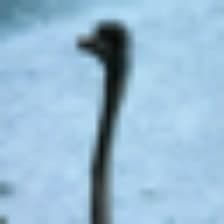}
      \caption{RGB image.}
      \label{fig_2:sub1}
    \end{subfigure}
    \begin{subfigure}[t]{.35\textwidth}
      \centering
      \includegraphics[width=\linewidth]{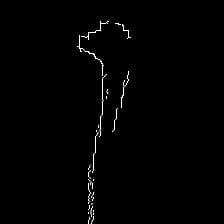}
      \caption{Image with edge detection.}
      \label{fig_2:sub2}
    \end{subfigure}
    \caption{A class sample from the CIFAR dataset~\cite{cifar10}. In (a) the input is a $224 \times 224$ RGB image, used as the models' first input stream. In (b), the same image with edge detection is the input of the second stream.}
    \label{fig_2}
\end{figure}

Three sets of data were used to evaluate our fusion strategy: CIFAR-10~\cite{cifar10}, CIFAR-50, and CIFAR-100~\cite{cifar100}. Three different sizes of datasets were chosen to check our strategy performance with a distinct amount of data samples. In order to create images for the second stream input, we created a mirrored input using an edge extraction created with a Canny edge detector~\cite{Canny:1986:CAE:11274.11275}. Figure~\ref{fig_2} presents an example of our input streams. Also, it is important to emphasize that all images were resized to a $224 \times 224$ shape with the purpose that they fit most of our networks default input shapes.

\begin{itemize}
    \item The CIFAR-10 dataset is composed of 60000 colored images in a 32x32 resolution, distributed in 10 balanced classes. There are 50000 training images and 10000 test images.
    \item Instead of 10 classes, like the previously cited CIFAR-10, the CIFAR-100 has 100 classes with 600 32x32 images in each class. Inside the one hundred classes, there are 20 superclasses that generalize certain labels, but these were not considered in this work.
    \item To produce the CIFAR-50 dataset, we randomly chose fifty classes from the original CIFAR-100 dataset. The final samples consisted of 50 classes with 600 images each, 500 for training and 100 for testing.
\end{itemize}

\subsection{NORB Object Recognition Dataset}
The main goal of this dataset is to recognize 3D objects from shapes. It has pictures of 50 toys belonging to 5 general categories: airplanes, four-legged animals, trucks, human figures, and cars. The dataset provides a pair of images for every toy: the items were imaged by a pair of cameras under 9 elevations (30 to 70 degrees every 5 degrees), 6 lighting conditions, and 18 azimuths (0 to 340 every 20 degrees)~\cite{norb}.

\begin{figure}[!htpb]
  \centering
  \includegraphics[width=.5\textwidth]{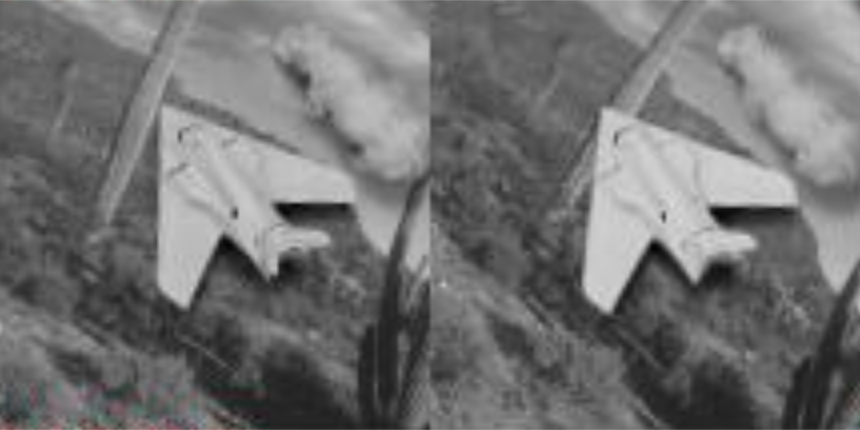}
  \caption{Pair of images found in NORB dataset.}
  \label{norb}
\end{figure}

\subsection{Hardware and training time}
All networks -- except for AlexNet -- were adapted from their previous implementation on the Keras library~\cite{chollet2015keras}. Fusion operations were created using tensor operations available in the Tensorflow framework~\cite{abadi2016tensorflow}. The experiments were trained in one RTX 3090 GPU and all experiments, including the NORB dataset, took approximately 4 months.

\section{Results and discussion}
\label{results}
All of the CIFAR datasets had their training set subsampled in a cross-validation procedure with 5 folds. The following accuracy results are presented by fold. No data augmentation nor transfer learning was used in the whole process. Due to a large number of experiments, we chose to gather all accuracies in one figure. In all plots, the color \textit{blue} will represent the late fusion version of the network. Colors \textit{orange, green}, and \textit{red} are the L-fusion architectures with average, sum, and subtraction operations.

\begin{figure}
     \centering
     \begin{subfigure}[b]{0.3\textwidth}
         \centering
         \includegraphics[width=\textwidth]{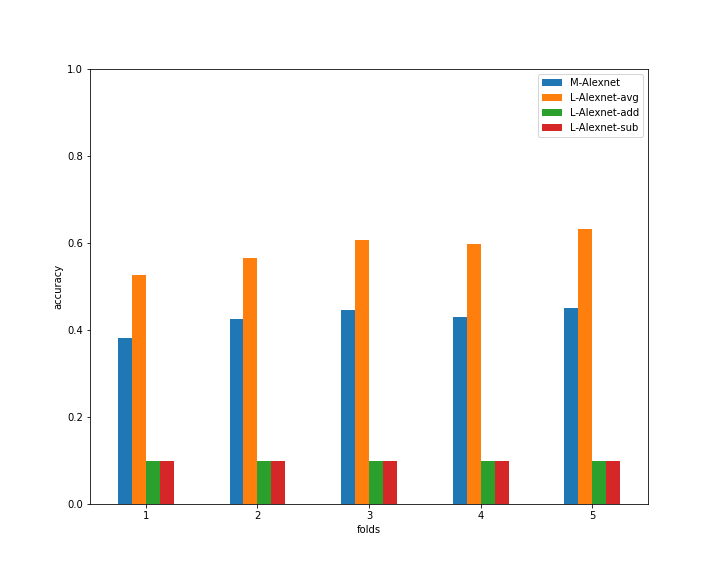}
         \caption{Alexnet - CIFAR-10}
     \end{subfigure}
     \hfill
     \begin{subfigure}[b]{0.3\textwidth}
         \centering
         \includegraphics[width=\textwidth]{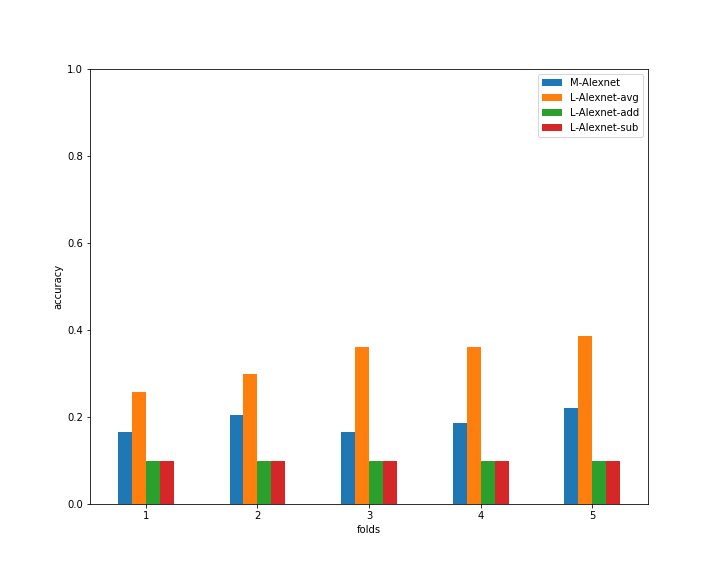}
         \caption{Alexnet - CIFAR-50}
     \end{subfigure}
     \hfill
     \begin{subfigure}[b]{0.3\textwidth}
         \centering
         \includegraphics[width=\textwidth]{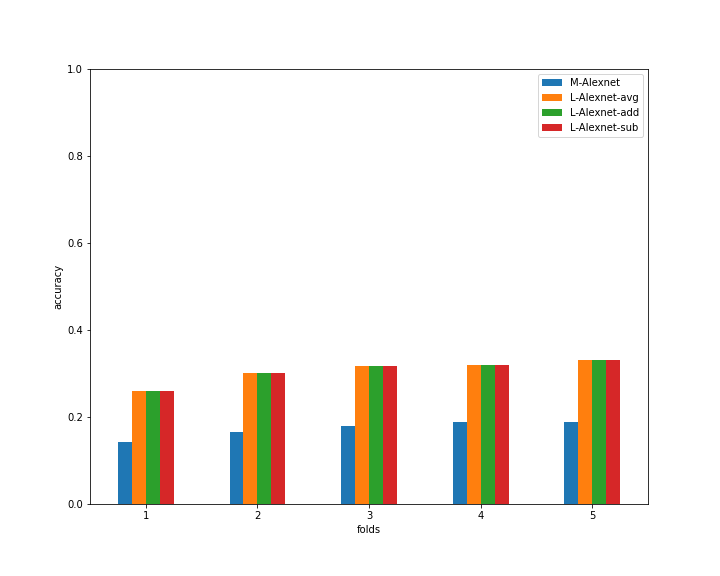}
         \caption{Alexnet - CIFAR-100}
     \end{subfigure}
     \begin{subfigure}[b]{0.3\textwidth}
         \centering
         \includegraphics[width=\textwidth]{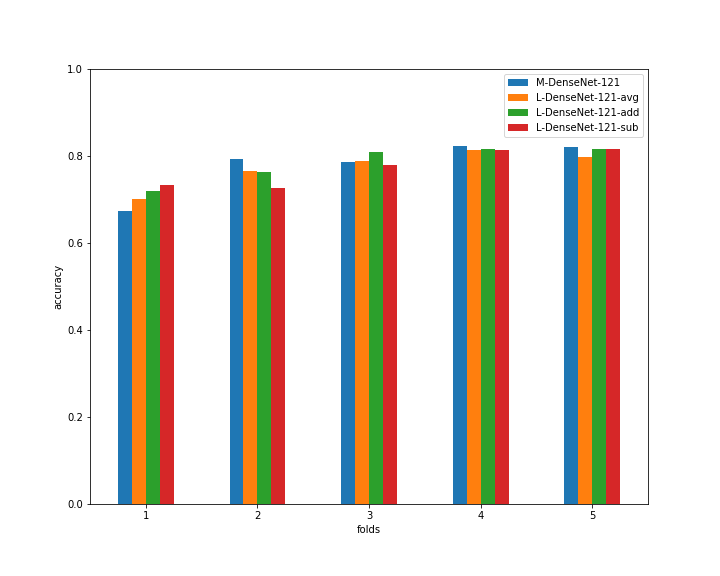}
         \caption{DenseNet-121 - CIFAR-10}
     \end{subfigure}
     \hfill
     \begin{subfigure}[b]{0.3\textwidth}
         \centering
         \includegraphics[width=\textwidth]{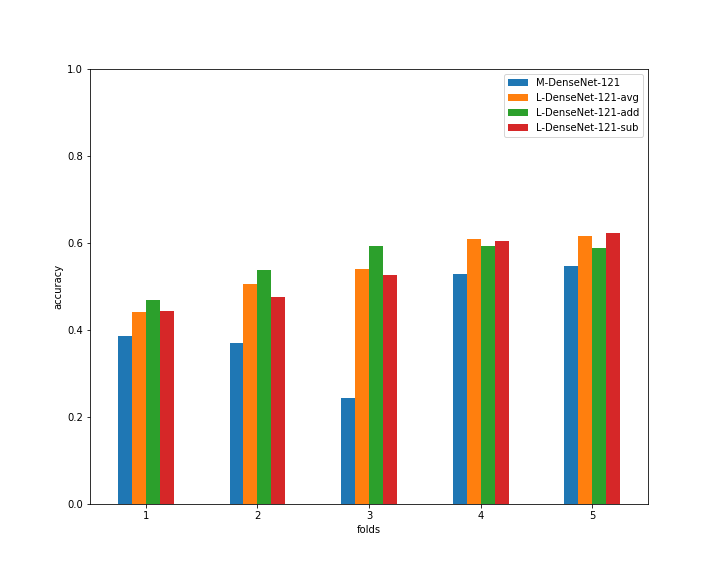}
         \caption{DenseNet-121 - CIFAR-50}
     \end{subfigure}
     \hfill
     \begin{subfigure}[b]{0.3\textwidth}
         \centering
         \includegraphics[width=\textwidth]{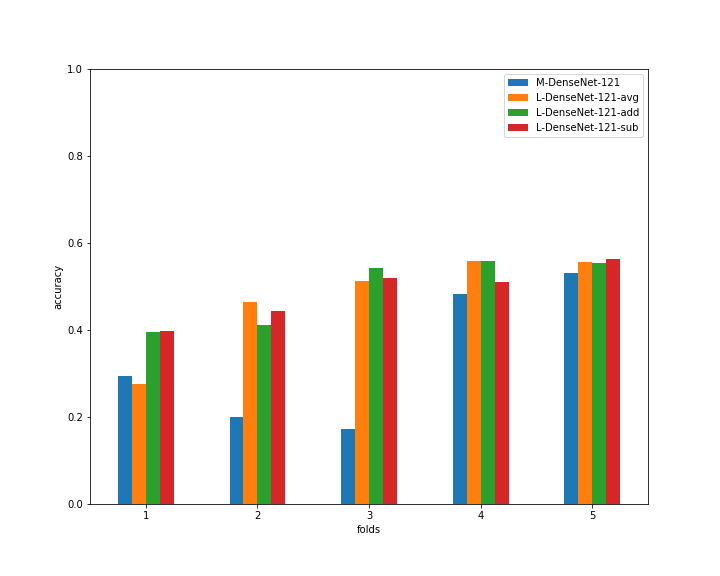}
         \caption{DenseNet-121 - CIFAR-100}
     \end{subfigure}
    \begin{subfigure}[b]{0.3\textwidth}
         \centering
         \includegraphics[width=\textwidth]{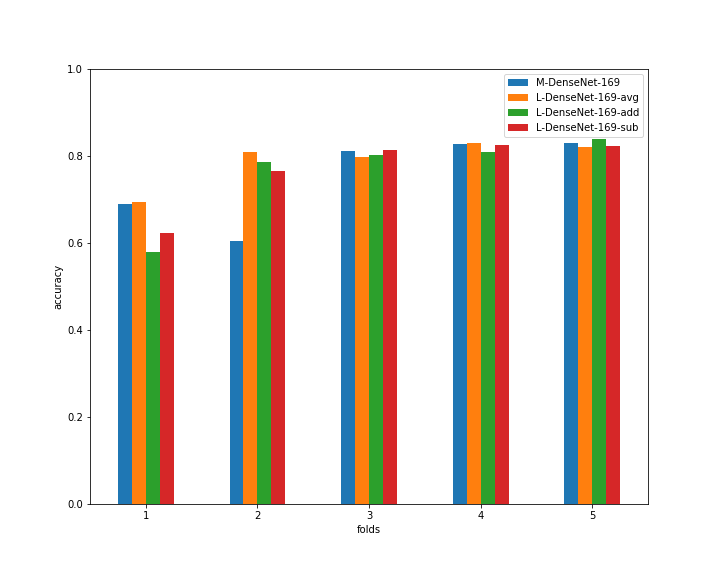}
         \caption{DenseNet-169 - CIFAR-10}
     \end{subfigure}
     \hfill
     \begin{subfigure}[b]{0.3\textwidth}
         \centering
         \includegraphics[width=\textwidth]{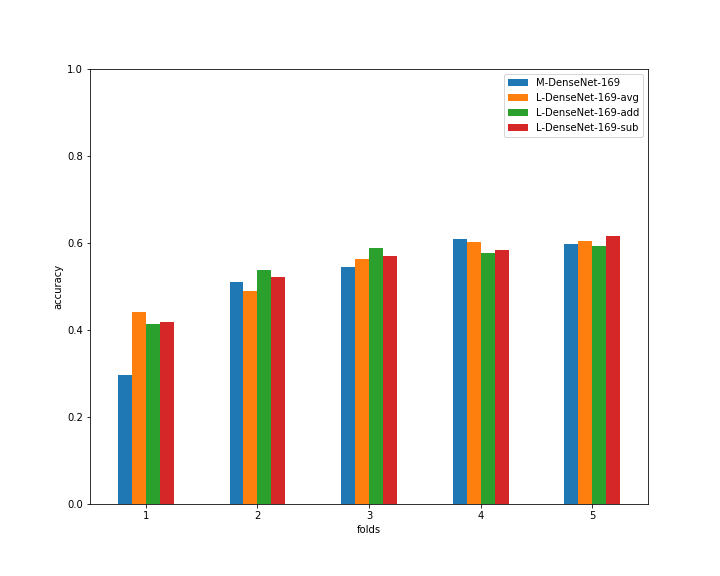}
         \caption{DenseNet-169 - CIFAR-50}
     \end{subfigure}
     \hfill
     \begin{subfigure}[b]{0.3\textwidth}
         \centering
         \includegraphics[width=\textwidth]{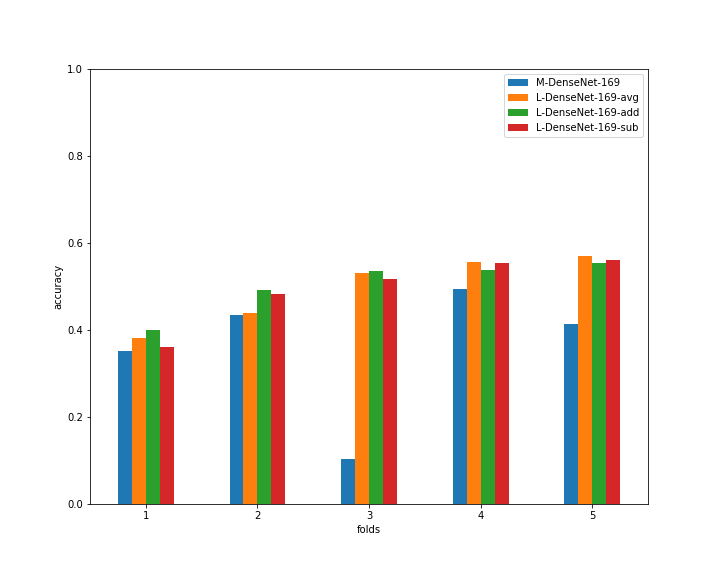}
         \caption{DenseNet-169 - CIFAR-100}
     \end{subfigure}
     \begin{subfigure}[b]{0.3\textwidth}
         \centering
         \includegraphics[width=\textwidth]{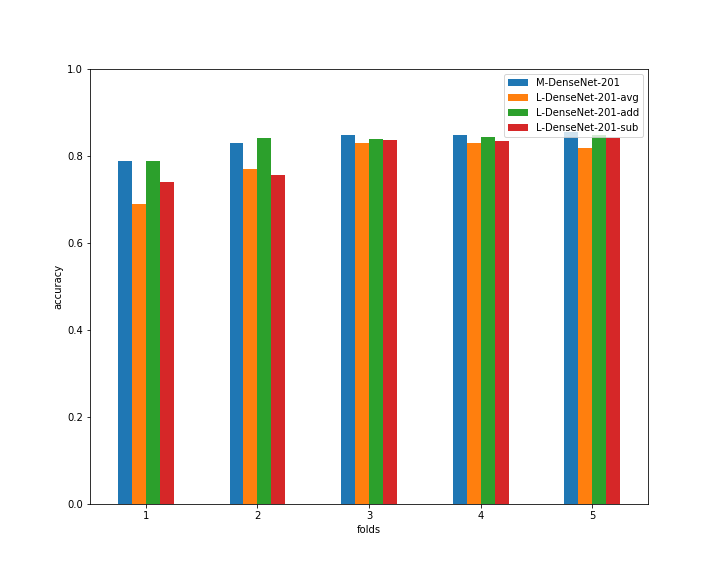}
         \caption{DenseNet-201 - CIFAR-10}
     \end{subfigure}
     \hfill
     \begin{subfigure}[b]{0.3\textwidth}
         \centering
         \includegraphics[width=\textwidth]{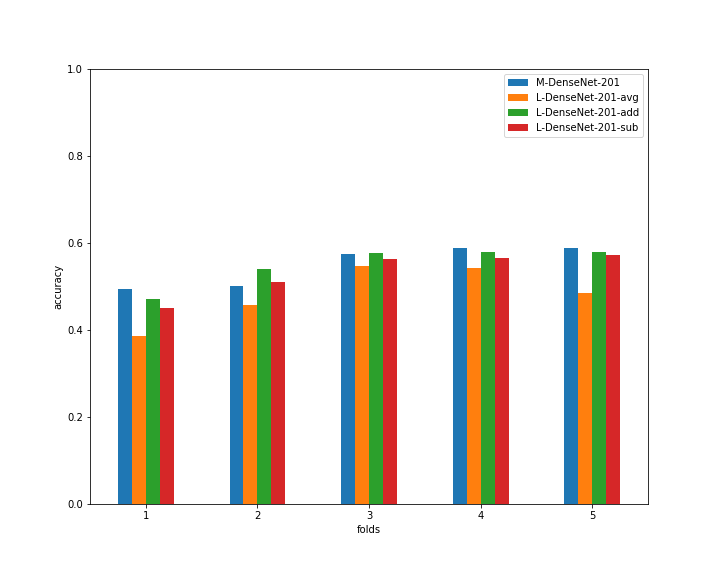}
         \caption{DenseNet-201 - CIFAR-50}
     \end{subfigure}
     \hfill
     \begin{subfigure}[b]{0.3\textwidth}
         \centering
         \includegraphics[width=\textwidth]{dense201-100.png}
         \caption{DenseNet-201 - CIFAR-100}
     \end{subfigure}
        \caption{Compilation of accuracy comparisons using Alexnet and DenseNet architecture variations.}
        \label{fig:alexdense}
\end{figure}

\begin{figure}
     \centering
     \begin{subfigure}[b]{0.3\textwidth}
         \centering
         \includegraphics[width=\textwidth]{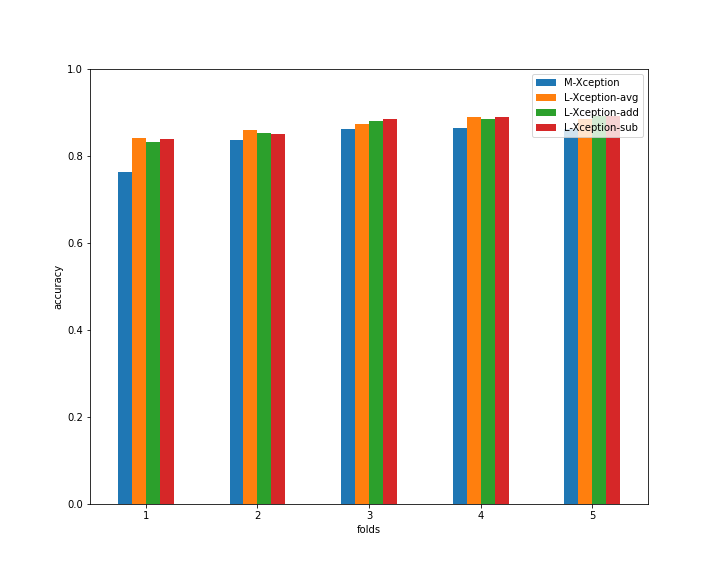}
         \caption{Xception - CIFAR-10}
     \end{subfigure}
     \hfill
     \begin{subfigure}[b]{0.3\textwidth}
         \centering
         \includegraphics[width=\textwidth]{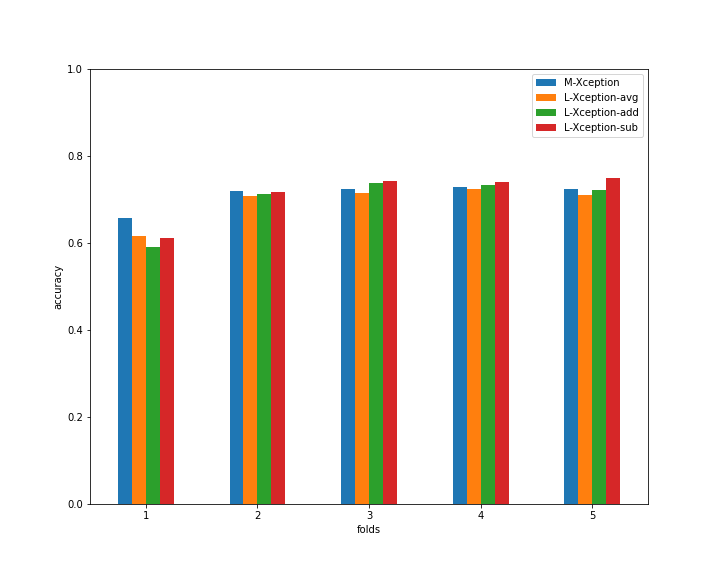}
         \caption{Xception - CIFAR-50}
     \end{subfigure}
     \hfill
     \begin{subfigure}[b]{0.3\textwidth}
         \centering
         \includegraphics[width=\textwidth]{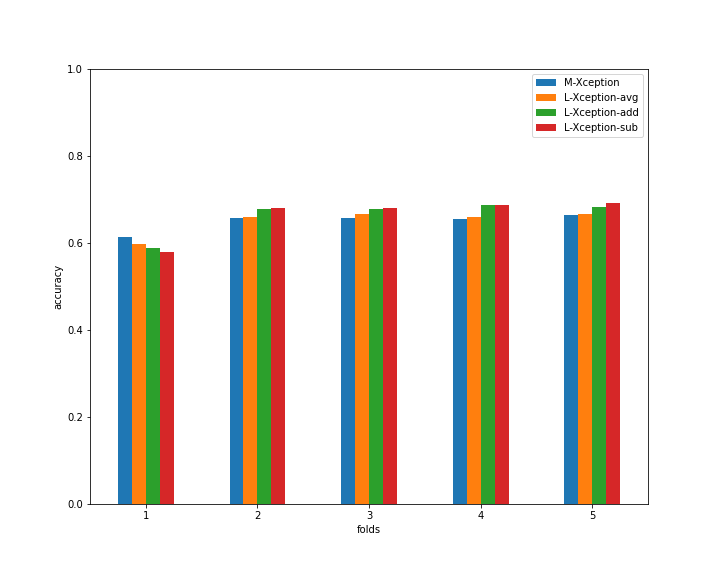}
         \caption{Xception - CIFAR-100}
     \end{subfigure}
     \begin{subfigure}[b]{0.3\textwidth}
         \centering
         \includegraphics[width=\textwidth]{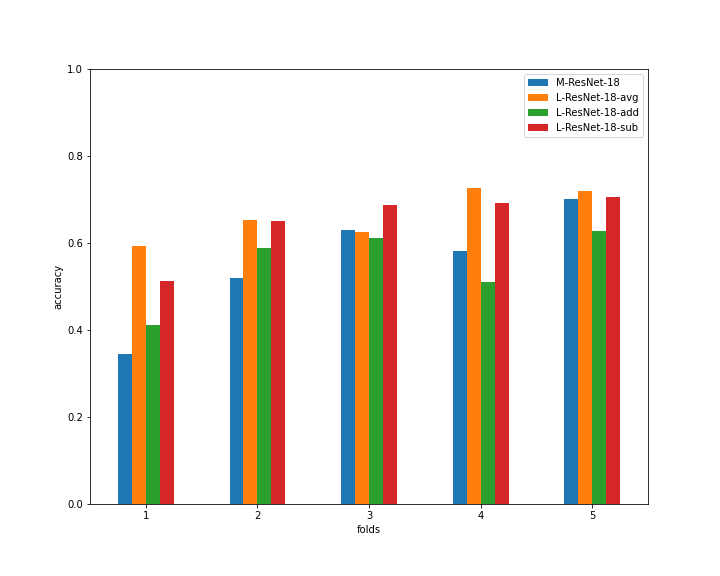}
         \caption{ResNet-18 - CIFAR-10}
     \end{subfigure}
     \hfill
     \begin{subfigure}[b]{0.3\textwidth}
         \centering
         \includegraphics[width=\textwidth]{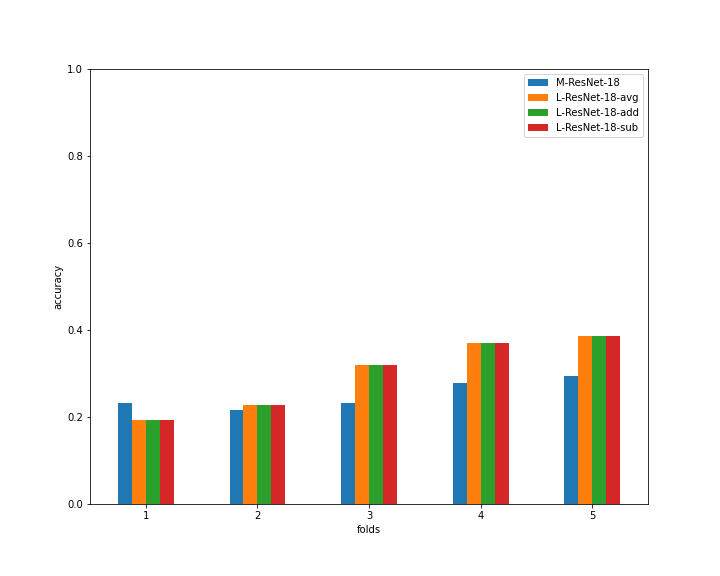}
         \caption{ResNet-18 - CIFAR-50}
     \end{subfigure}
     \hfill
     \begin{subfigure}[b]{0.3\textwidth}
         \centering
         \includegraphics[width=\textwidth]{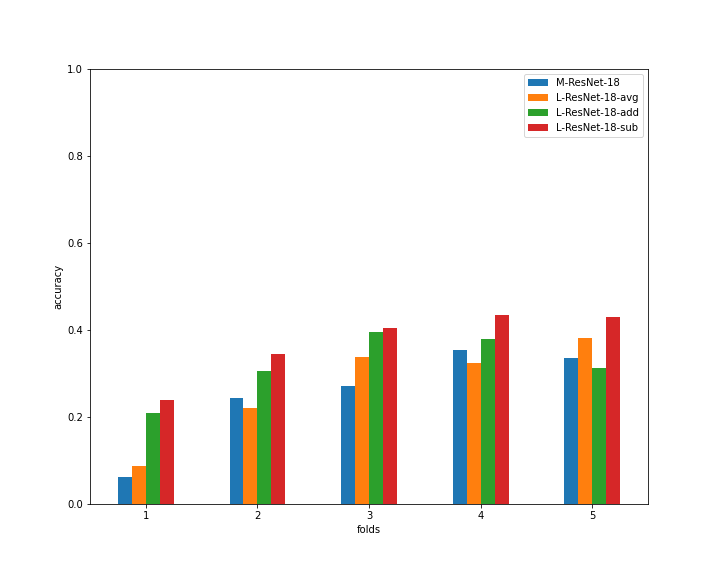}
         \caption{ResNet-18 - CIFAR-100}
     \end{subfigure}
    \begin{subfigure}[b]{0.3\textwidth}
         \centering
         \includegraphics[width=\textwidth]{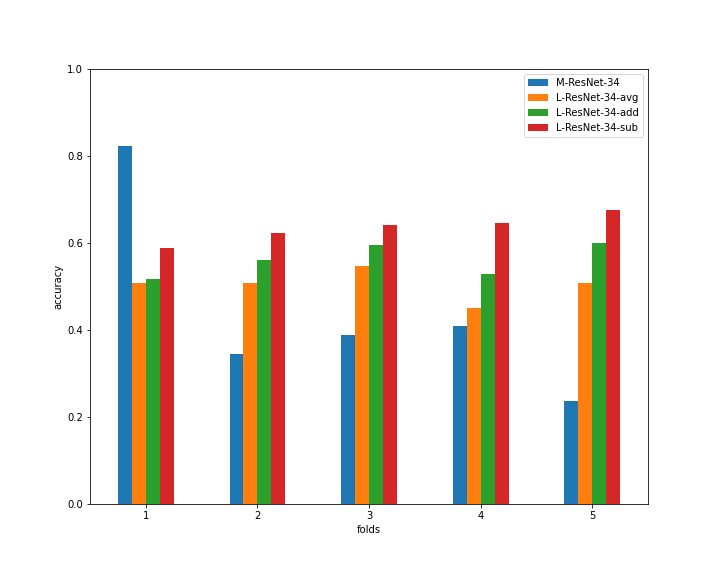}
         \caption{ResNet-34 - CIFAR-10}
     \end{subfigure}
     \hfill
     \begin{subfigure}[b]{0.3\textwidth}
         \centering
         \includegraphics[width=\textwidth]{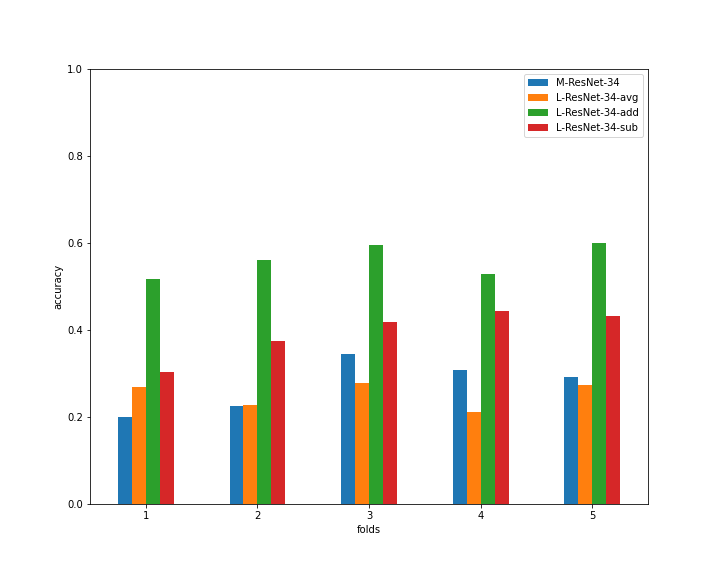}
         \caption{ResNet-34 - CIFAR-50}
     \end{subfigure}
     \hfill
     \begin{subfigure}[b]{0.3\textwidth}
         \centering
         \includegraphics[width=\textwidth]{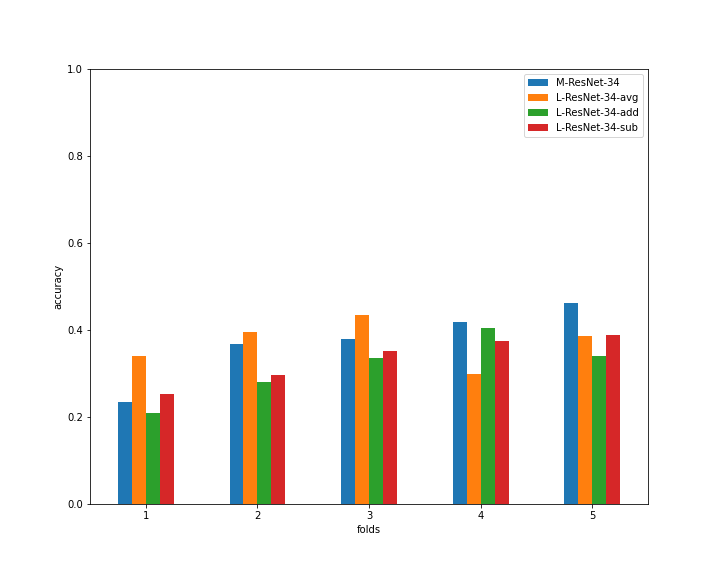}
         \caption{ResNet-34 - CIFAR-100}
     \end{subfigure}
     \begin{subfigure}[b]{0.3\textwidth}
         \centering
         \includegraphics[width=\textwidth]{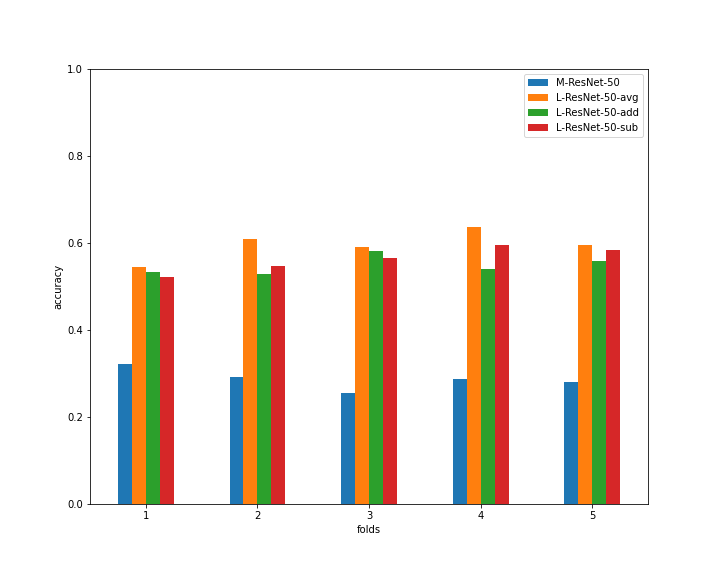}
         \caption{ResNet-50 - CIFAR-10}
     \end{subfigure}
     \hfill
     \begin{subfigure}[b]{0.3\textwidth}
         \centering
         \includegraphics[width=\textwidth]{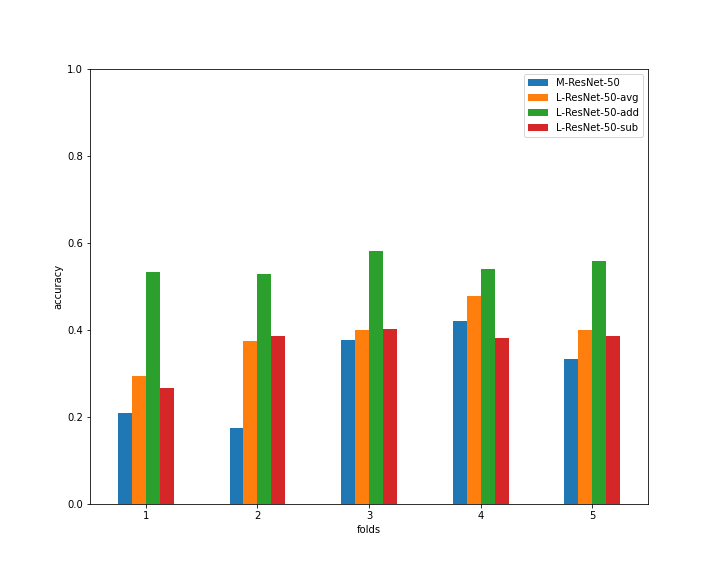}
         \caption{ResNet-50 - CIFAR-50}
     \end{subfigure}
     \hfill
     \begin{subfigure}[b]{0.3\textwidth}
         \centering
         \includegraphics[width=\textwidth]{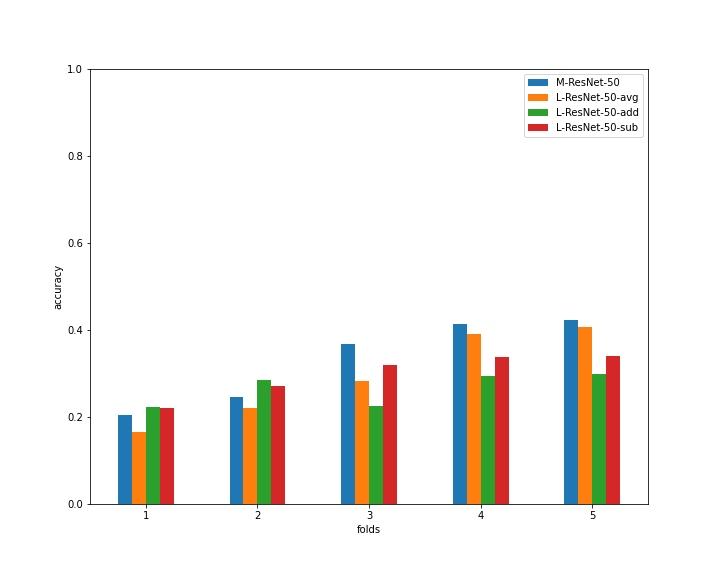}
         \caption{ResNet-50 - CIFAR-100}
     \end{subfigure}
        \caption{Compilation of accuracy comparisons using Xception and ResNet architecture variations.}
        \label{fig:xcepres}
\end{figure}

\begin{figure}
    \centering
     \begin{subfigure}[b]{0.49\textwidth}
         \centering
         \includegraphics[width=\textwidth]{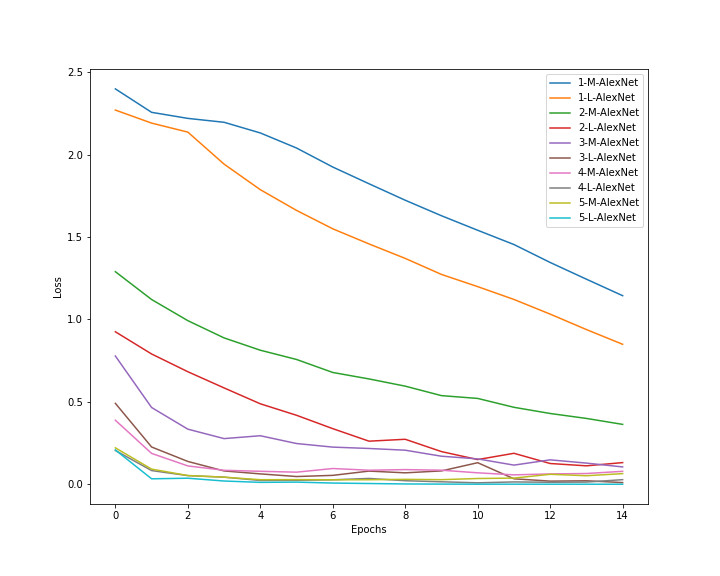}
         \caption{AlexNet losses per fold - CIFAR-10 dataset}
     \end{subfigure}
     \hfill
     \begin{subfigure}[b]{0.49\textwidth}
         \centering
         \includegraphics[width=\textwidth]{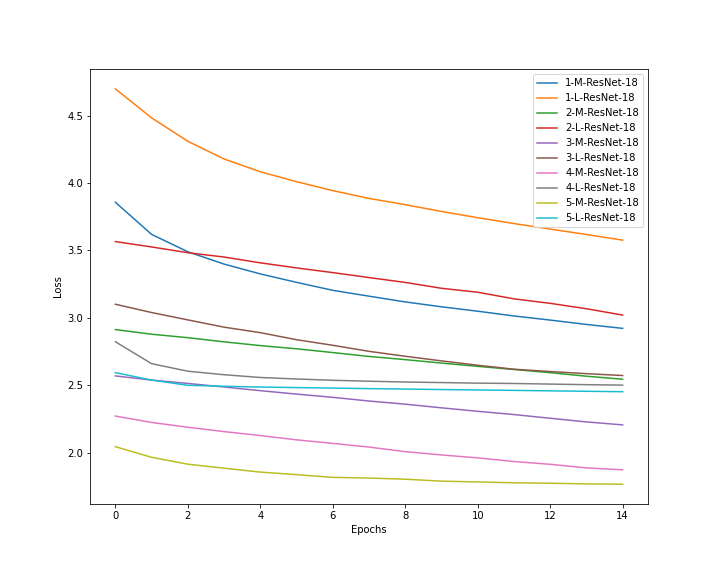}
         \caption{ResNet-18 losses per fold - CIFAR-10 dataset}
     \end{subfigure}
          \begin{subfigure}[b]{0.5\textwidth}
         \centering
         \includegraphics[width=\textwidth]{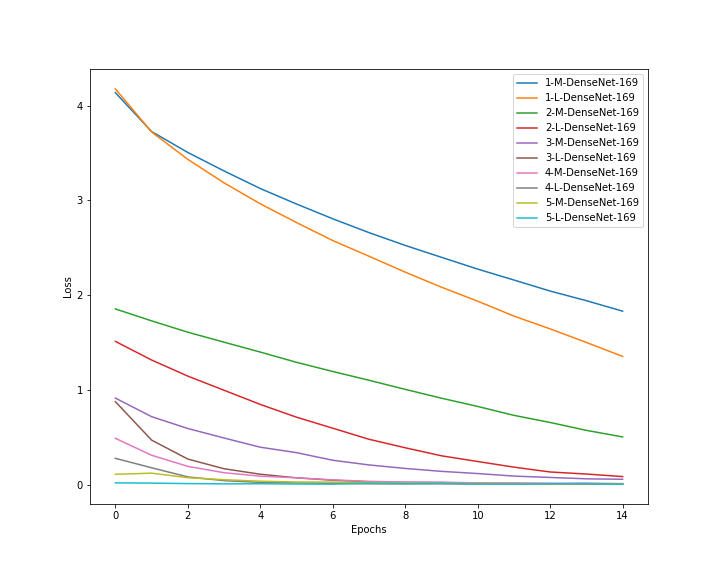}
         \caption{DenseNet-169 losses per fold - CIFAR-100 dataset}
     \end{subfigure}
        \caption{Loss comparison per fold showing different architectures and distinct inputs.}
        \label{fig:losses}
\end{figure}

Considering the increase of the network parameters presented in Table~\ref{parameter}, but also an accuracy gaining of $5.66\%$ using a default L-strategy average operation over a simple late fusion technique, independently of the given dataset or dataset quality, it is fair to say that the occasional increase in parameters is outweighed by the performance boost. Our losses graph, Figure~\ref{fig:losses}, also shows that the lattice-adapted networks tend to converge faster.

The goal of this work is not to beat any kind of state-of-the-art models, as previously noted. Thus, it is remarkable to achieve strong accuracies with very degraded inputs. Our L-Xception with the average operation, for example, reached an $87.02\%$ accuracy on CIFAR-10.

Also, using the same dataset as a comparator, a late fusion Alexnet went from $42.69\%$ to $58.66\%$. Alexnet has a very straightforward backbone, being easy to implement and it can run smoothly on plain hardware.

Tables~\ref{tb:allacc10},~\ref{tb:allacc50}, and~\ref{tb:allacc100} present the mean accuracy of the folds for all models and datasets tested here, being a summarized version of the graphs presented in Figures~\ref{fig:alexdense}, and~\ref{fig:xcepres}.

\newpage
\KOMAoptions{paper=landscape,pagesize}
\newgeometry{top=2cm,textwidth=19cm,textheight=18cm}

\begin{table}[!htpb]
\centering
\begin{tabular}{c|c|c|c|c|c|c|c|c|c}
\multicolumn{2}{c|}{Architecture}                    & Alexnet          & DenseNet-121     & DenseNet-169 & DenseNet-201     & Xception         & ResNet-18        & ResNet-34        & ResNet-50         \\ \hline\hline
\multicolumn{2}{c|}{MCNN}       & 0.4269           & 0.77938          & 0.75268      & \textbf{0.83398} & 0.83796          & 0.55552          & 0.44064          & 0.28746             \\ \hline
\multirow{3}{*}{LCNN} & \textit{avg}     & \textbf{0.58664} & 0.54228          & 0.54044      & 0.57288          & 0.8702           & \textbf{0.66346} & 0.50456          & \textbf{0.59532}\\ \cline{2-10} 
                                       & \textit{add}    & 0.1              & \textbf{0.78486} & 0.76316      & 0.83204          & 0.8689           & 0.5501           & 0.56054          & 0.54876             \\ \cline{2-10} 
                                       & \textit{sub} & 0.1              & 0.774            & 0.7706       & 0.80192          & \textbf{0.87172} & 0.64964          & \textbf{0.63494} & 0.56318
\end{tabular}
\caption{Mean accuracies of all trained models for CIFAR-10.}
\label{tb:allacc10}
\end{table}

\begin{table}[!htpb]
\centering
\begin{tabular}{c|c|c|c|c|c|c|c|c|c}
\multicolumn{2}{c|}{Architecture}                    & Alexnet          & DenseNet-121     & DenseNet-169 & DenseNet-201     & Xception         & ResNet-18        & ResNet-34        & ResNet-50         \\ \hline\hline
\multicolumn{2}{c|}{MCNN}       & 0.18856         & 0.3574          & 0.35958          & 0.54936          & 0.71128         & 0.2506           & 0.27376          & 0.30304             \\ \hline
\multirow{3}{*}{LCNN} & \textit{avg}     & \textbf{0.336} & 0.47392          & 0.49604          & 0.48392          & 0.69468         & 0.29904          & 0.25228          & \textbf{0.3896}\\ \cline{2-10} 
                                       & \textit{add}    & 0.1             & \textbf{0.55664} & 0.5418           & \textbf{0.60304} & 0.69968         & 0.37636          & 0.37844          & 0.34484\\ \cline{2-10} 
                                       & \textit{sub} & 0.1             & 0.53472          & \textbf{0.54212} & 0.576            & \textbf{0.7122} & \textbf{0.39008} & \textbf{0.39488} & 0.36508
\end{tabular}
\caption{Mean accuracies of all trained models for CIFAR-50.}
\label{tb:allacc50}
\end{table}

\begin{table}[!htpb]
\centering
\begin{tabular}{c|c|c|c|c|c|c|c|c|c}
\multicolumn{2}{c|}{Architecture}                    & Alexnet          & DenseNet-121     & DenseNet-169 & DenseNet-201     & Xception         & ResNet-18        & ResNet-34        & ResNet-50         \\ \hline\hline
\multicolumn{2}{c|}{MCNN}       & 0.17308         & \textbf{0.77938} & 0.75268         & \textbf{0.83398} & 0.64986          & 0.25368         & \textbf{0.3718} & \textbf{0.3094}\\ \hline
\multirow{3}{*}{LCNN} & \textit{avg}     & \textbf{0.3058} & 0.77306          & \textbf{0.7905} & 0.78772          & 0.65028          & 0.27036         & 0.37116         & 0.29326\\ \cline{2-10} 
                                       & \textit{add}    & 0.1             & 0.49234          & 0.5043          & 0.54928          & \textbf{0.66294} & 0.32088         & 0.31384         & 0.2655\\ \cline{2-10} 
                                       & \textit{sub} & 0.1             & 0.48726          & 0.49514         & 0.53238          & 0.66398          & \textbf{0.3709} & 0.3244         & 0.29738
\end{tabular}
\caption{Mean accuracies of all trained models for CIFAR-100.}
\label{tb:allacc100}
\end{table}

\newpage
\KOMAoptions{paper=portrait,pagesize}
\recalctypearea


After the experiments with variant and degraded CIFAR datasets, we decided to understand our proposal better by applying our method in a real and not overused dataset. Table~\ref{tb:norb} presents a comparison between three different-sized architectures in three forms: single stream, multistream with late fusion, and multistream with our lattice cross-connection. Results show that at least one lattice operation could outperform all other results, demonstrating the flexibility of the technique when we come across not-so-good results. 

\begin{table}[]
\centering
\begin{tabular}{c|c|c|c|c}
\multicolumn{2}{c|}{Architecture}                    & DenseNet-169    & ResNet-50       & Alexnet         \\ \hline\hline
\multicolumn{2}{c|}{Single stream - left image}      & 0.6021          & 0.2             & 0.2             \\ \hline
\multicolumn{2}{c|}{Multistream - late fusion}       & 0.3727          & 0.6785          & 0.2             \\ \hline
\multirow{3}{*}{Multistream - lattice} & \textit{average}     & \textbf{0.7091} & 0.4682          & 0.2             \\ \cline{2-5} 
                                       & \textit{addition}    & 0.5253          & 0.3138          & 0.2             \\ \cline{2-5} 
                                       & \textit{subtraction} & 0.5156          & \textbf{0.7558} & \textbf{0.8321}
\end{tabular}
\caption{NORB accuracies on the test set.}
\label{tb:norb}
\end{table}


\subsection{What if L-strategy doesn't improve my results?}
There is a possibility that the described module ant its default operators will over-amplify signals in a way that the model will simply not learn. Using signal processing knowledge, we know that when this kind of peaks occurs, we can perform a compression to approximate low and high signals. Therefore, we propose the use of a logarithm-based compression function that will balance the broad range, reducing the difference between large and small values. This function is described in Equation~\ref{eq:log}, where $s$ and $\bar{s}$ are the incoming signal streams.

\begin{equation}
\centering
    F(s,\bar{s}) = s - \log(1 - \bar{s}).
\label{eq:log}
\end{equation}

\begin{figure}[H]
  \centering
  \includegraphics[width=.5\textwidth]{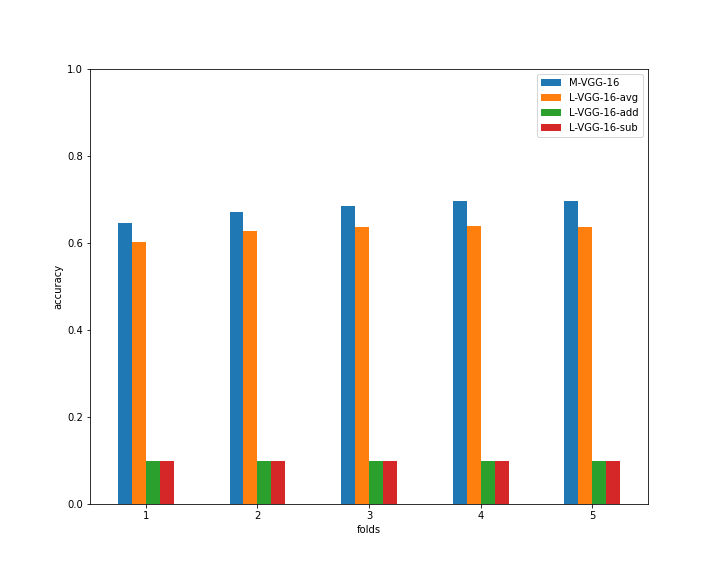}
  \caption{VGG-16 results on CIFAR-10.}
  \label{fig:vgg}
\end{figure}

An example is our L-VGG-16~\cite{vgg} implementation. Our first run using an \textit{average} function went well, but not enough to beat the M-VGG-16 architecture. Further runs with other defined operators were boosting both tensor streams and our model found its local minimum very fast, without room for optimizations. Figure~\ref{fig:vgg} shows a comparison between operations and modeling type.

Table~\ref{tb:vgg} presents all obtained VGG-16 accuracies for each fold on test set. We can clearly see that our log-compression function improved the overall results.

\begin{table*}[!htpb]
\centering
\begin{tabular}{c|ccccc}
\multicolumn{1}{c|}{}                       & \multicolumn{5}{c}{Architecture}                                                                                                                                                            \\ \cline{2-6} 
\multicolumn{1}{c|}{\multirow{-2}{*}{Fold}}  & M-VGG-16 & L-VGG-16-avg & L-VGG-16-log     & L-VGG-16-add & L-VGG-16-sub \\\hline\hline
1       & 0.6454   & 0.6021       & \textbf{0.7265}  & 0.1          & 0.1          \\
2       & 0.672    & 0.628        & \textbf{0.7426}  & 0.1          & 0.1          \\
3       & 0.6847   & 0.638        & \textbf{0.7562}  & 0.1          & 0.1          \\
4       & 0.6963   & 0.6397       & \textbf{0.7594}  & 0.1          & 0.1          \\
5       & 0.6966   & 0.6377       & \textbf{0.7555}  & 0.1          & 0.1          \\\hline
Average & 0.679    & 0.6291       & \textbf{0.74804} & 0.1          & 0.1         
\end{tabular}
\caption{VGG-16 accuracies on the test set for each fold.}
\label{tb:vgg}
\end{table*}

Notice that the aforementioned technique was meant to be used with all default hyperparameters, still, it does not work for all situations. A VGG-19 with standard initialization will continue to have its signal overamplified, converging too quickly. However, basic optimization approaches are sufficient to smoothly train the network: the learning rate, for example, controls the adaptability of the dataset to the network and it affects directly in weights updated, hence, when we are faced with large amplifications caused by the L-strategy, it is usually enough to decrease the starting learning rate or apply schedulers. To illustrate this, we present in Table~\ref{tb:vgg19} a comparison between a multistream late fusion approach and the L-strategy approach with the non-default learning rate of $10^{-4}$, forasmuch as the regular parameterization generated unusable results.

\begin{table*}[!htpb]
\centering
\begin{tabular}{c|cccc}
\multicolumn{1}{c|}{}                       & \multicolumn{4}{c}{Architecture}                                                                                                                                                            \\ \cline{2-5} 
\multicolumn{1}{c|}{\multirow{-2}{*}{Fold}}  & M-VGG-19 & L-VGG-19-avg & L-VGG-19-add & L-VGG-19-sub \\\hline\hline
1       & 0.6663   &    0.5643          & 0.6511       & \textbf{0.6671}       \\
2       & 0.6625   &    0.5064          & 0.6952       & \textbf{0.7052}       \\
3       & 0.6932   &    0.5252          & 0.7121       & \textbf{0.7184}       \\
4       & 0.6911   &    0.5485          & \textbf{0.7289}       & 0.7245       \\
5       & 0.6455   &    0.5553          & 0.7338       & \textbf{0.7349}       \\ \hline
Average & 0.67172  &    0.53994          & 0.70422      & \textbf{0.71002}
\end{tabular}
\caption{VGG-19 accuracies on the test set for each fold.}
\label{tb:vgg19}
\end{table*}

Finally, we can observe that the presented results in this Section are not linear, there isn't an absolute better fusion operation. Although we recommend on using \textit{average}, we displayed our results using a trial-and-error method.

\section{Conclusions and future works}
\label{conclusion}
In this work, we presented a novel fusion technique in order to improve multistream image classifications. Using different backbones and 4 datasets, experimental results show that the proposed strategy has faster convergence and also has a flexibility of operational switches. Likewise, the proposed fusion demonstrated robustness and stability, even when distractors are used as inputs.

Still, there is no definitive technique. We showed that we can obtain an increase of accuracy up to $63.21\%$, as well as our strategy overamplifies so much the signal that the default hypermeters are not enough to train a good model. Our goal to reuse previous state-of-the-art architectures with few modifications to keep them in the game was successfully achieved. In other words: the L-strategy module can make older models usable for brand-new challenges.

Future work will focus on the inclusion of new steps exploring improvement categories: the next move is to improve a range of other models, following an exploratory line of research for new techniques to give an extra life to models becoming obsolete. Additionally, we intend to explore not only CNNs, but every kind of structure that includes an ReLU activation, such as Long short-term memory (LSTM) and Generative Adversarial Networks (GAN), evaluation the proposed technique in other data modalities.

\bibliographystyle{elsarticle-num}  
\bibliography{references}

\end{document}